\pdfoutput=1
\documentclass[11pt]{article}

\usepackage[T1]{fontenc}
\usepackage[utf8]{inputenc}
\usepackage{lmodern}
\usepackage[margin=1in]{geometry}
\usepackage{amsmath,amssymb,amsthm,mathtools}
\usepackage{booktabs}
\usepackage{array}
\usepackage{tabularx}
\usepackage{enumitem}
\usepackage{xcolor}
\usepackage{fancyvrb}
\usepackage[section]{placeins}
\usepackage[numbers,sort&compress]{natbib}
\usepackage[hidelinks]{hyperref}
\usepackage{url}
\usepackage[nameinlink]{cleveref}

\newtheorem{theorem}{Theorem}
\newtheorem{lemma}[theorem]{Lemma}

\theoremstyle{definition}

\theoremstyle{remark}
\newtheorem{remark}[theorem]{Remark}

\newcommand{\R}{\mathbb{R}}
\newcommand{\E}{\mathbb{E}}
\newcommand{\Prob}{\mathbb{P}}
\newcommand{\Id}{\mathbf{I}}
\newcommand{\Kmat}{\mathbf{K}}
\newcommand{\Rowcov}{\boldsymbol{\Sigma}}
\newcommand{\op}{\mathrm{op}}
\newcommand{\lmin}{\lambda_{\min}}
\newcommand{\lmax}{\lambda_{\max}}
\newcommand{\Loewner}{\preceq}
\DeclareUrlCommand\lean{\urlstyle{tt}}

\DefineVerbatimEnvironment{leanlisting}{Verbatim}{fontsize=\footnotesize,frame=single,framesep=3pt}
\DeclareUnicodeCharacter{00D7}{\ensuremath{\times}}
\DeclareUnicodeCharacter{03A0}{\ensuremath{\Pi}}
\DeclareUnicodeCharacter{03A9}{\ensuremath{\Omega}}
\DeclareUnicodeCharacter{03B4}{\ensuremath{\delta}}
\DeclareUnicodeCharacter{03B8}{\ensuremath{\theta}}
\DeclareUnicodeCharacter{03B9}{\ensuremath{\iota}}
\DeclareUnicodeCharacter{03BB}{\ensuremath{\lambda}}
\DeclareUnicodeCharacter{03BC}{\ensuremath{\mu}}
\DeclareUnicodeCharacter{03BD}{\ensuremath{\nu}}
\DeclareUnicodeCharacter{03C1}{\ensuremath{\rho}}
\DeclareUnicodeCharacter{03C3}{\ensuremath{\sigma}}
\DeclareUnicodeCharacter{03C9}{\ensuremath{\omega}}
\DeclareUnicodeCharacter{1D34}{\textsuperscript{H}}
\DeclareUnicodeCharacter{1D65}{\ensuremath{_v}}
\DeclareUnicodeCharacter{2016}{\ensuremath{\|}}
\DeclareUnicodeCharacter{2022}{\ensuremath{\bullet}}
\DeclareUnicodeCharacter{2080}{\ensuremath{_0}}
\DeclareUnicodeCharacter{2081}{\ensuremath{_1}}
\DeclareUnicodeCharacter{2082}{\ensuremath{_2}}
\DeclareUnicodeCharacter{2102}{\ensuremath{\mathbb{C}}}
\DeclareUnicodeCharacter{2115}{\ensuremath{\mathbb{N}}}
\DeclareUnicodeCharacter{211D}{\ensuremath{\mathbb{R}}}
\DeclareUnicodeCharacter{2192}{\ensuremath{\to}}
\DeclareUnicodeCharacter{2200}{\ensuremath{\forall}}
\DeclareUnicodeCharacter{2202}{\ensuremath{\partial}}
\DeclareUnicodeCharacter{2203}{\ensuremath{\exists}}
\DeclareUnicodeCharacter{2208}{\ensuremath{\in}}
\DeclareUnicodeCharacter{2211}{\ensuremath{\sum}}
\DeclareUnicodeCharacter{2227}{\ensuremath{\wedge}}
\DeclareUnicodeCharacter{222B}{\ensuremath{\int}}
\DeclareUnicodeCharacter{2260}{\ensuremath{\neq}}
\DeclareUnicodeCharacter{2264}{\ensuremath{\leq}}
\DeclareUnicodeCharacter{2265}{\ensuremath{\geq}}

\newcommand{\codeurl}{\url{https://github.com/allenhaozhu/NTK-Lean}}

\title{Verifying and Strengthening Neural Tangent Kernel Theory%
\thanks{Accepted to KDD 2027 (Research Track). Lean~4 artifact: \codeurl.}}
\author{Hao Zhu\\
CSIRO Technology\\
\texttt{allenhaozhu@gmail.com}}
\date{}

\begin{document}
\maketitle

\begin{abstract}
Neural tangent kernel (NTK) theory explains why very wide networks trained by gradient methods converge in the lazy-training regime. Its standard proof composes matrix concentration, random Gram matrices, continuous-time dynamics, spectral perturbation and a bootstrap argument, and it has so far been checked only informally. We present a Lean~4 formalisation of this argument for two-layer networks and an end-to-end, machine-checked theorem: for a smooth bounded activation (we instantiate $\sigma=\tanh$ in Lean), frozen biases, paired $\pm1$ initialisation and explicit width conditions, with probability at least $1-\delta$ the gradient flow exists, is unique, and the training residual decays as $\|r(t)\|^2\le\|y\|^2e^{-\lambda_0t/2}$. The only data assumption is the standard non-degeneracy condition $\lambda_0>0$ on the population NTK; initial coercivity, kernel drift, the movement bootstrap, the residual ODE and existence and uniqueness of the flow are proved. The formalisation also sharpened the informal analysis in two places. First, the matrix-Bernstein carrier is parametric in its variance proxy, which exposed that the standard proxy ($R^2$ times the number of independent summands) is loose for NTK Gram matrices; replacing it with the (classical) row-covariance proxy gives, numerically for $\tanh$ with the data held fixed, estimated width certificates for the concentration step that are about $36$--$53\times$ smaller at $(m,d)=(16,8)$ and $94$--$121\times$ smaller at $(32,16)$; Lean certifies the inequality, the ratios are numerical. Second, the Hessian of the network has operator norm $\Theta(\sqrt{p})$ on the sup-norm parameter ball (a paper calculation), so the sup-norm Taylor route cannot give the expected $O(1/\sqrt{p})$ linearisation error; an $L^2$ parameter geometry does. The required widths are polynomial and far larger than practical ones, so the practical reach of the result is narrow; we discuss this limitation explicitly.
\end{abstract}

\section{Introduction}\label{sec:intro}

Learning theory increasingly depends on long chains of results from probability, matrix analysis, optimisation and differential equations. When a guarantee is reused (for example to justify a width, a sample size or a step size), every hypothesis, quantifier and constant in that chain must compose, and informal review is an unreliable check of this. A proof assistant such as Lean~4~\cite{demoura2021lean4} with Mathlib~\cite{mathlib2024} accepts a theorem only when each step type-checks against the definitions it uses.

The neural tangent kernel (NTK) lazy-training argument~\cite{jacot2018neural,chizat2018lazy,du2019gradient,li2018learning,allenzhu2019convergence,zou2020gradient,arora2019finegrained} is a good test case because it is a composition rather than a single lemma: matrix Bernstein concentration of a random Gram matrix, a lower bound on the smallest eigenvalue of the initial NTK, a residual ODE obtained from gradient-flow dynamics, a bootstrap that keeps the kernel close to its initial value, and a Gr\"onwall step. Existing Lean formalisations of learning theory~\cite{zhang2026ai4slt,sonoda2025rademacher} cover empirical processes and Rademacher complexity, not this convergence argument.

\paragraph{Contributions.}
\begin{enumerate}[leftmargin=1.5em,itemsep=2pt,topsep=2pt]
\item \textbf{An end-to-end machine-checked NTK convergence theorem} (\Cref{sec:main}). For the two-layer network with $1/\sqrt{p}$ scaling, frozen biases and paired $\pm1$ initialisation, with probability at least $1-\delta$ over the initial weights the gradient flow exists, is unique on every horizon, and the residual decays exponentially. The data assumption $\lambda_0>0$ and explicit width conditions are the only non-structural hypotheses (\Cref{sec:assumptions}); the movement/drift bootstrap that earlier versions of this work took as a hypothesis is now derived. The theorem is instantiated in Lean for $\sigma=\tanh$.
\item \textbf{A sharper variance proxy for the concentration step} (\Cref{sec:rowcov}). The verified matrix-Bernstein carrier takes its variance proxy as an argument. Using the row-covariance proxy $\|\sum_i\E X_i^2\|$ in place of the standard proxy ($R^2$ times the number of summands) is a classical refinement~\cite[\S6.1]{tropp2015introduction}, and we do not claim it as new mathematics. What the formalisation contributed is to make the proxy an explicit interface, which exposed the slack, and to let the sharper proxy be used without re-proving Bernstein. On data-conditional $\tanh$ instances the resulting estimated concentration-step width certificate is $36$--$53\times$ smaller at $(m,d)=(16,8)$ and $94$--$121\times$ smaller at $(32,16)$ (medians $45\times$ and $103\times$; numerical, \Cref{tab:width}).
\item \textbf{A norm-dependence observation for linearisation} (\Cref{sec:norm}). The Hessian of the network has operator norm $\Theta(\sqrt{p})$ on the sup-norm parameter ball, so the Taylor-remainder \emph{proof route} in the sup norm cannot yield $O(1/\sqrt{p})$; in an $L^2$ geometry it does. The lower bound is a paper calculation; the $L^2$ bound is formalised as a conditional lemma.
\end{enumerate}
All Lean theorems cited are \texttt{sorry}-free, and their axioms are among the three standard ones ($\mathsf{propext}$, $\mathsf{Classical.choice}$, $\mathsf{Quot.sound}$). \Cref{tab:claims} maps every claim to its Lean declaration, and \Cref{app:listings} reproduces the statements of the main theorems.

\section{Setting and the target theorem}\label{sec:setting}

\paragraph{Network.} We study the two-layer network
\[
f_\theta(x)=\frac{1}{\sqrt p}\sum_{j=1}^{p}a_j\,\sigma\bigl(\langle w_j,x\rangle+b_j\bigr),\qquad x\in\R^d,
\]
with trainable parameters $\theta=(a,w)$, frozen biases $b$, and activation $\sigma$. The running example throughout the paper is $\sigma=\tanh$. Training inputs are $x_1,\dots,x_m$ with labels $y\in\R^m$; in Lean the number of inputs is \lean{n} and the width is \lean{k + k}. The loss is $L(\theta)=\tfrac12\sum_i(f_\theta(x_i)-y_i)^2$, the residual is $r(\theta)=(f_\theta(x_i)-y_i)_i$, and gradient flow is $\dot\theta=-\nabla L(\theta)$.

\paragraph{Initialisation.} We use the paired initialisation: $k=p/2$ neurons $(w_i,b_i)$, $i=1,\dots,k$, are drawn i.i.d.\ from a law $\nu$, each is duplicated, and the two copies get output weights $+1$ and $-1$. Then $f_{\theta_0}\equiv0$ and $r(0)=-y$.

\paragraph{NTK.} The NTK Gram matrix at $\theta$ is $\Kmat(\theta)_{ab}=\langle\nabla_\theta f_\theta(x_a),\nabla_\theta f_\theta(x_b)\rangle$. At initialisation each neuron contributes $\frac1p A$ with
\[
A_{ab}=\sigma(z_{a})\sigma(z_{b})+\sigma'(z_{a})\sigma'(z_{b})\langle x_a,x_b\rangle,\qquad z_{a}=\langle w,x_a\rangle+b,
\]
where we used $a^2=1$. The two copies of a pair have $a=\pm1$ and the same $(w,b)$, hence the same matrix; so with $A_i$ the matrix of pair $i$,
\[
\Kmat(\theta_0)=\frac1p\sum_{j=1}^{p}A_j=\frac1k\sum_{i=1}^{k}A_i,
\]
an average of $k=p/2$ \emph{independent} matrices (not $p$). Along gradient flow the residual satisfies the \emph{residual ODE} $\dot r=-\Kmat(\theta(t))\,r$.

\paragraph{The concentration step in words.} The argument first shows that $\Kmat(\theta_0)$ is close to its expectation over the random neurons, $\E\Kmat(\theta_0)$, the population NTK. Write $\Kmat(\theta_0)-\E\Kmat(\theta_0)=\sum_{i=1}^kX_i$ with $X_i=\frac1k(A_i-\E A_i)$. The $X_i$ are independent, centred, symmetric random matrices (one per neuron pair), and each pair's contribution is bounded, $\|X_i\|_\op\le R$. Matrix Bernstein~\cite{tropp2012user,tropp2015introduction} then gives, for every $t>0$,
\begin{equation}\label{eq:bernstein}
\Prob\Bigl(\Bigl\|\sum_{i=1}^k X_i\Bigr\|_\op\ge t\Bigr)\le 4m\exp\Bigl(-\frac{t^2}{2v+2Rt/3}\Bigr),\qquad v\ \ge\ \Bigl\|\sum_{i=1}^k\E X_i^2\Bigr\|_\op .
\end{equation}
Here we use the constant of the Lean statement: \lean{Matrix.bernstein_full} bounds the one-sided tail $\Prob(\lmax(\sum_iX_i)\ge t)$ by $2m\exp(\cdot)$, and applying it to $\pm X_i$ with a union bound gives $4m$; textbook two-sided versions~\cite{tropp2015introduction} have $2m$. If $\lmin(\E\Kmat(\theta_0))\ge\lambda_0$, then on the complementary event $\lmin(\Kmat(\theta_0))\ge\lambda_0-t$. We call a bound $\lmin(\Kmat(\theta_0))\ge\rho>0$ \emph{initial coercivity}; it is simply a lower bound on the smallest eigenvalue of the initial NTK.

\paragraph{The informal target.} The classical statement~\cite{du2019gradient,arora2019finegrained} is: if the initial NTK has $\lmin\ge\rho$ and the kernel stays within $\rho/2$ of its initial value along the flow, then $\|r(t)\|^2\le\|r(0)\|^2e^{-\rho t}$. An earlier version of our development formalised exactly this \emph{conditional} statement, with coercivity, drift and the residual ODE as hypotheses. The theorem in \Cref{sec:main} replaces all three hypotheses by proofs.

\section{Assumptions}\label{sec:assumptions}

\Cref{tab:assumptions} lists every hypothesis of the final theorem, whether it is assumed or proved, and how restrictive it is.

\begin{table}[!htbp]
\centering
\caption{Assumptions of the end-to-end theorem (\Cref{thm:main}). ``Assumed'' items are hypotheses of the Lean statement; ``proved'' items are Lean theorems used in its proof.}
\label{tab:assumptions}
\footnotesize
\begin{tabularx}{\linewidth}{@{}>{\raggedright\arraybackslash}p{3.1cm}>{\raggedright\arraybackslash}X>{\raggedright\arraybackslash}p{1.6cm}>{\raggedright\arraybackslash}X@{}}
\toprule
Item & Meaning & Status & Restrictiveness \\
\midrule
\multicolumn{4}{@{}l}{\emph{Data}}\\
$\lambda_0>0$ & $\lmin$ of the population NTK $\E_\nu A$ on the fixed inputs is at least $\lambda_0$ (Lean: \lean{hpop}). & assumed & Standard non-degeneracy condition, but not automatic: for an arbitrary $\nu$ it can fail (e.g.\ for $\nu$ a point mass at $w=0$, $b=0$ and $\sigma=\tanh$, $\tanh(0)=0$ and $\tanh'(0)=1$ give $(\E_\nu A)_{ab}=\langle x_a,x_b\rangle$, the input Gram matrix, which is singular whenever $m>d$, as in our instances). Known sufficient conditions: Gaussian weights $w\sim\mathcal N(0,I_d)$ with zero bias and pairwise non-parallel inputs, for ReLU~\cite[Thm.~3.1]{du2019gradient} and for analytic non-polynomial activations such as $\tanh$~\cite{du2019deep}; these concern the $\sigma'\sigma'\langle x,x'\rangle$ block, which $\E_\nu A$ dominates. Kept as an assumption; not proved in Lean. Estimated numerically (quadrature; not a certificate), for $\tanh$, Gaussian $w$, zero bias and inputs uniform on the sphere: $\lambda_0\in[0.048,0.066]$ at $(d,m)=(8,16)$ and $[0.054,0.086]$ at $(16,32)$ (10 input draws). \\
\midrule
\multicolumn{4}{@{}l}{\emph{Setup}}\\
Activation & $\sigma$ differentiable, $|\sigma|\le M$, $|\sigma'|\le M'$, $\sigma'$ Lipschitz. & assumed; proved for $\tanh$ & Covers $\tanh$, sigmoid. Excludes ReLU and clipped ReLU. For $\tanh$ ($M=M'=1$, $\mathrm{Lip}(\sigma')\le1$) the hypotheses are proved in Lean. \\
Biases & frozen during training & assumed & Structural simplification. \\
Initialisation & paired $\pm1$ output weights, so $r(0)=-y$; neurons i.i.d.\ from any law $\nu$ & assumed & Standard symmetrisation trick; removes the random initial output. \\
Loss, scaling & squared loss, $1/\sqrt p$ output scaling, bounded inputs $\|x_i\|\le X$ & assumed & Standard NTK parameterisation. \\
\midrule
\multicolumn{4}{@{}l}{\emph{Width}}\\
Concentration & two matrix-Bernstein width bounds (\lean{ntkBernsteinWidth}), $p=\tilde O(m^2\lambda_0^{-2}\log(m/\delta))$ & assumed (explicit) & For $\tanh$ at the $\lambda_0$ above: $p\gtrsim1.8\cdot10^8$--$2.7\cdot10^8$ ($m=16$) and $3.9\cdot10^8$--$9.0\cdot10^8$ ($m=32$). \\
Drift & $p\ge C\,m^3\|y\|^2/\lambda_0^4$ (\lean{gfMovementBound}, \lean{gfDriftConst}); $\sim m^4/\lambda_0^4$ when $\|y\|^2\sim m$ & assumed (explicit) & The restrictive part: same polynomial regime as~\cite{du2019gradient}. For $\tanh$: $p\gtrsim2.5\cdot10^{15}$--$5.4\cdot10^{15}$ ($m=16$), $1.0\cdot10^{16}$--$5.5\cdot10^{16}$ ($m=32$). Far beyond practical widths; reflects the known gap between lazy training and practice~\cite{chizat2018lazy}. \\
\midrule
\multicolumn{4}{@{}l}{\emph{Proved, not assumed}}\\
\multicolumn{4}{@{}>{\raggedright\arraybackslash}p{\dimexpr\linewidth\relax}@{}}{Initial coercivity with probability $\ge1-\delta$ (including Bernstein's measurability and integrability side conditions); measurability of the good event; the flow field equals $-\nabla L$; the residual ODE along the flow; kernel drift and per-neuron movement along the flow (bootstrap); existence and uniqueness of the gradient flow on every $[0,T]$.}\\
\bottomrule
\end{tabularx}
\end{table}

The concrete widths in \Cref{tab:assumptions} are computed by \texttt{calibration/widths\_tanh.py} from the closed-form Lean definitions with $M=M'=\mathrm{Lip}(\sigma')=1$, $X=1$, $\|y\|=\sqrt m$, $\delta=0.01$, at the minimum and median of the ten numerical $\lambda_0$ values. They are the widths of the formal theorem and should not be confused with the concentration-only certificates of \Cref{sec:rowcov}, which concern a single step at a fixed deviation $t$.

\section{The machine-checked convergence theorem}\label{sec:main}

\begin{theorem}[End-to-end lazy training]\label{thm:main}
Let $\sigma$ be differentiable with $|\sigma|\le M$, $|\sigma'|\le M'$ and $\sigma'$ $L_{\sigma'}$-Lipschitz; let $\|x_i\|\le X$; let the neurons of the paired initialisation be i.i.d.\ from $\nu$, and assume $\lmin(\E_\nu\Kmat(\theta_0))\ge\lambda_0>0$. If the width satisfies the explicit concentration and drift conditions of \Cref{tab:assumptions}, then there is a measurable event $G$ with $\Prob(G)\ge1-\delta$ such that for every initialisation in $G$ and every horizon $T$:
\begin{enumerate}[label=(\roman*),itemsep=1pt,topsep=2pt]
\item a gradient-flow trajectory on $[0,T]$ starting at $\theta_0$ exists;
\item it is unique on $[0,T]$ (among trajectories with the stated right derivative);
\item every such trajectory satisfies $\|r(t)\|^2\le\|y\|^2e^{-\lambda_0t/2}$ for all $t\in[0,T]$.
\end{enumerate}
Lean: \lean{fullNet_gradient_flow_lazy_training_whp_exists_uniform}. For $\sigma=\tanh$ the activation hypotheses are discharged in Lean (\lean{fullNet_gradient_flow_lazy_training_whp_exists_uniform_tanh}).
\end{theorem}

The full Lean statement is reproduced verbatim in \Cref{app:listings}. Its proof composes the following verified components.

\paragraph{Initial coercivity with high probability.} Matrix Bernstein (\lean{Matrix.bernstein_full}, proved from Lieb's concavity theorem~\cite{lieb1973convex} through Tropp's argument~\cite{tropp2012user}) is applied separately to the $\sigma\sigma$ block and the $\sigma'\sigma'\langle x,x'\rangle$ block, each at deviation $\lambda_0/4$, so that $\lmin(\Kmat(\theta_0))\ge\lambda_0/2$ with probability $\ge1-\delta$ (\lean{fullTrainingKernel_antisymInit_coercive_whp}). Its measurability and integrability side conditions are discharged for continuous bounded $\sigma,\sigma'$, and the good event is shown to be closed, hence measurable.

\paragraph{Residual ODE.} The flow field is proved to be exactly $-\nabla L$ (\lean{fderiv_sqLossFullNet_apply}), and along any trajectory of that field the residual satisfies $\dot r=-\Kmat(\theta(t))r$, with $\Kmat$ the NTK Gram matrix of parameter gradients (\lean{residualFullNet_ode_on_Icc}).

\paragraph{Bootstrap.} Kernel stability needs bounded movement of the parameters, and bounded movement needs the residual decay that kernel stability provides. \lean{fullNet_gradient_flow_lazy_training} closes this loop for any gradient-flow trajectory by a first-exit argument: if $|a_j(0)|\le1$, $\lmin(\Kmat(\theta_0))\ge\rho$ and the width conditions hold, then on $[0,T]$ the residual satisfies $\|r(t)\|^2\le\|r(0)\|^2e^{-\rho t}$ and each neuron moves by at most $\frac{2}{\rho}\cdot\frac{(M+2M'X)\sqrt m\,\|r(0)\|}{\sqrt p}=O\bigl(\sqrt m\|r(0)\|/(\rho\sqrt p)\bigr)$. No drift or movement hypothesis is used.

\paragraph{Existence and uniqueness.} The flow is constructed by Picard--Lindel\"of for a retracted, globally Lipschitz field on a ball and shown to stay inside the ball by the movement bound; uniqueness follows by Gr\"onwall (\lean{exists_fullNet_gradient_flow}, \lean{fullNet_gradient_flow_unique}).

\begin{remark}[Correction to the submitted version]
In the version submitted for review, the per-neuron movement bound (the hypothesis \lean{hmove} of the drift interface \lean{fullNetwork_lazy_kernel_drift_parametric}) was supplied as a hypothesis, and some listings in the paper did not match the code. The movement bound is now derived by \lean{fullNet_gradient_flow_lazy_training}. The older conditional interfaces (\lean{fullNetwork_lazy_kernel_drift_parametric}, \lean{ntk_lazy_training_carrier_parametric}) remain in the artifact as reusable parametric statements and are documented there as superseded; they are not on the path of \Cref{thm:main}. All listings in this paper are extracted from the source by a script.
\end{remark}

\begin{table}[!htbp]
\centering
\caption{Claim-to-Lean map. Files (in parentheses) are relative to \texttt{NTK/MathlibExt/} and omit \texttt{.lean}; ``P'' abbreviates \texttt{Probability}, ``PT.'' the namespace \texttt{ProbabilityTheory}, and ``NMP.'' the namespace \texttt{NTK.MathlibExt.Probability}.}
\label{tab:claims}
\footnotesize
\begin{tabularx}{\linewidth}{@{}>{\raggedright\arraybackslash}p{4.6cm}>{\raggedright\arraybackslash}X@{}}
\toprule
Claim & Lean declaration (file) \\
\midrule
\Cref{thm:main} (all $T$, w.p.\ $\ge1-\delta$) & \lean{PT.fullNet_gradient_flow_lazy_training_whp_exists_uniform} (\lean{P/NTKLazyTrainingWhpExists}) \\
\Cref{thm:main} at $\sigma=\tanh$ & \lean{PT.fullNet_gradient_flow_lazy_training_whp_exists_uniform_tanh} (\lean{P/NTKTanhInstance}) \\
$\tanh'=1-\tanh^2$; $\tanh'$ is 1-Lipschitz; $|\tanh''|\le1$ & \lean{NTK.Tanh.hasDerivAt_tanh}, \lean{NTK.Tanh.lipschitzWith_deriv_tanh}, \lean{NTK.Tanh.abs_iteratedDeriv_two_tanh_le_one} (\lean{P/NTKTanhInstance}) \\
Fixed horizon $T$ & \lean{PT.fullNet_gradient_flow_lazy_training_whp_exists} (\lean{P/NTKLazyTrainingWhpExists}) \\
Good event is measurable & \lean{PT.measurableSet_setOf_antisymInit_coercive} (\lean{P/NTKLazyTrainingWhpExists}) \\
Bootstrap: decay and movement along any flow & \lean{PT.fullNet_gradient_flow_lazy_training} (\lean{P/NTKGradientFlowBootstrap}) \\
Existence and uniqueness of the flow & \lean{PT.exists_fullNet_gradient_flow}, \lean{PT.fullNet_gradient_flow_unique} (\lean{P/NTKGradientFlowExistence}) \\
Initial coercivity w.h.p. & \lean{PT.fullTrainingKernel_antisymInit_coercive_whp} (\lean{P/NTKInitCoercivityWhp}) \\
Flow field $=-\nabla L$ & \lean{PT.fderiv_sqLossFullNet_apply} (\lean{P/NTKGradientFlowResidual}) \\
Residual ODE & \lean{PT.residualFullNet_ode_on_Icc} (\lean{P/NTKGradientFlowResidual}) \\
$r(0)=-y$ under paired initialisation & \lean{PT.residualFullNet_antisymInit} (\lean{P/NTKAntisymmetricInit}) \\
Bernstein side conditions & \lean{PT.empiricalFullNTK_matrix_bernstein_real_of_continuous} (\lean{P/NTKBernsteinRegularity}) \\
Lieb's concavity theorem (trace-exp form) & \lean{Matrix.lieb_tropp_concave} (\lean{MatrixAnalysis/LiebTroppConcave}) \\
Matrix Bernstein & \lean{Matrix.bernstein_full} (\lean{MatrixAnalysis/MatrixBernsteinFinal}) \\
Row-covariance bridge (\Cref{lem:bridge}) & \lean{Matrix.row_covariance_loewner_bound} (\lean{MatrixAnalysis/RowCovarianceLoewner}) \\
Row-covariance Bernstein (\Cref{thm:tightened}) & \lean{PT.ntk_matrix_bernstein_tightened} (\lean{P/NTKConcentrationMatrixBernsteinTightened}) \\
$L^2$ linearisation bound (conditional) & \lean{NMP.chizat_bach_analytical_L2_general} (\lean{Calculus/FullNetworkSecondFrechet}) \\
$L^2$ movement bound (conditional) & \lean{NMP.chizat_bach_energy_movement_L2_bounded_general} (\lean{P/ChizatBachEnergyMovementL2}) \\
Superseded conditional interfaces & \lean{NTK.ntk_lazy_training_carrier_parametric} (\lean{P/NTKLazyTrainingEndToEnd}), \lean{PT.fullNetwork_lazy_kernel_drift_parametric} (\lean{P/FullNTKLazyCarrier}) \\
\midrule
\multicolumn{2}{@{}>{\raggedright\arraybackslash}p{\linewidth}@{}}{\emph{Not in Lean:} $\kappa$ and width-certificate numerics (\Cref{sec:rowcov}); the $\Theta(\sqrt p)$ sup-norm calculation (\Cref{sec:norm}); $\lambda_0>0$ for a concrete $\nu$.}\\
\bottomrule
\end{tabularx}
\end{table}

\paragraph{What formalisation required.} Several steps that informal proofs leave implicit had to be built: Lieb's concavity theorem and the matrix Bernstein chain (absent from Mathlib at our pinned revision); an encoding of the NTK Gram matrix that the matrix and probability libraries both accept; existence and uniqueness of the flow, which informal arguments usually assume; the measurability of the good event; and the bootstrap, whose circularity informal proofs resolve by a continuity argument that had to be written out as a first-exit argument. The choice of norm on the parameter space, which informal proofs leave implicit, also had to be fixed; \Cref{sec:norm} shows that it matters.

\section{Row-covariance variance proxy and width certificates}\label{sec:rowcov}

This section is about the concentration step only. Its practical output is a width certificate: the width $p=2k$ at which \eqref{eq:bernstein} guarantees $\|\Kmat(\theta_0)-\E\Kmat(\theta_0)\|_\op<t$ with probability $1-\delta$. Because the proxy below is estimated numerically, the widths we report are \emph{estimated} certificate widths.

\subsection{What is proved}

The Lean carrier \lean{Matrix.bernstein_full} takes the variance proxy as an explicit argument $\sigma^2$ with $\sum_j\E X_j^2\Loewner\sigma^2\Id$. Like the textbook statement it also has measurability and integrability side conditions (measurability of $\lmax$ of the sum and integrability of the trace exponential); for the NTK summands they are discharged in Lean by \lean{empiricalFullNTK_matrix_bernstein_real_of_continuous}, for continuous bounded $\sigma,\sigma'$. Two choices of proxy are admissible.

\begin{lemma}[Row-covariance bridge; Lean: \lean{Matrix.row_covariance_loewner_bound}]\label{lem:bridge}
If $X_1,\dots,X_N$ are $n\times n$ complex matrices with $\|X_j\|_\op\le R$, then $\sum_jX_jX_j^*\Loewner R^2N\,\Id$.
\end{lemma}

\begin{theorem}[Row-covariance Bernstein; Lean: \lean{ntk_matrix_bernstein_tightened}]\label{thm:tightened}
Let $X_j$ be independent, centred, Hermitian, with $\|X_j\|_\op\le R$, and $\Rowcov=\sum_j\E X_jX_j^*\neq0$. Then for $t>0$,
$\Prob(\lmax(\sum_jX_j)\ge t)\le 2n\exp\bigl(-t^2/(2\|\Rowcov\|_\op+2Rt/3)\bigr)$, provided the measurability and integrability side conditions of \lean{Matrix.bernstein_full} hold (for the NTK summands they are discharged by \lean{empiricalFullNTK_matrix_bernstein_real_of_continuous}).
\end{theorem}

\Cref{lem:bridge} shows that the standard proxy $R^2N$ ($N$ the number of summands) always dominates $\|\Rowcov\|_\op$, so the row-covariance proxy is never worse. Both facts are classical~\cite{ahlswede2002strong,oliveira2010sums,tropp2012user,minsker2017some},~\cite[\S6.1]{tropp2015introduction}. Lean certifies the inequalities; the ratios below are numerical. \Cref{thm:main} currently uses the standard proxy; composing \Cref{thm:tightened} into it is future work.

\subsection{A width-independent normalisation}

To compare certificates consistently, work with the pair matrices $A_i$ of \Cref{sec:setting}, so that $\Kmat(\theta_0)=\frac1k\sum_{i=1}^kA_i$ with $k=p/2$ independent summands. For $\tanh$ and unit-norm inputs, $A_i$ is positive semidefinite with $\|A_i\|_\op\le\operatorname{tr}A_i\le2m$, so $\|A_i-\E A_i\|_\op\le R_A:=2m$. Let $v:=\|\E A_i^2\|_\op$; since $\E(A_i-\E A_i)^2\Loewner\E A_i^2$, this is a valid proxy. Both $R_A$ and $v$ are independent of $k$; applying \eqref{eq:bernstein} to $X_i=\frac1k(A_i-\E A_i)$ (range $R_A/k$, proxy $v/k$) and inverting gives the number of independent pairs and the total width
\begin{equation}\label{eq:pstar}
k^\star(v)=\frac{\log(4m/\delta)\,\bigl(2v+\tfrac{4}{3}mt\bigr)}{t^2},\qquad p^\star(v)=2k^\star(v).
\end{equation}
The standard proxy is $v_{\mathrm{std}}=R_A^2=4m^2$; the row-covariance proxy is $v=\|\E A_i^2\|_\op$, and we write $\kappa:=v_{\mathrm{std}}/v$. The ratio $p^\star_{\mathrm{std}}/p^\star$ does not depend on the constants $\log(4m/\delta)$ or the factor $2$.

\paragraph{Conditioning on the data.} The concentration is over the neurons with the inputs fixed, so the relevant proxy is the data-conditional $\Rowcov(X):=\E_w[A^2\mid X]$, not an average over inputs. We estimate $\Rowcov(X)$ by Monte Carlo ($4\cdot10^5$ weight samples) for ten independent draws of the inputs and report the distribution. A plug-in estimate of $v$ can underestimate the true proxy, and an underestimate makes the certificate too small. As a sensitivity check we also report a heuristic margin $v_+=\hat v+3\|\widehat{\mathrm{SE}}\|_F$, where $\widehat{\mathrm{SE}}$ is the batch-means standard error (20 batches) of each entry of the estimate of $\E_w[A^2\mid X]$ (motivated by $|\|M\|_\op-\|\hat M\|_\op|\le\|M-\hat M\|_F$); no coverage level is claimed for it. It exceeds $\hat v$ by at most $1.4\%$ at $(16,8)$ and $3.1\%$ at $(32,16)$. Setting: $\sigma=\tanh$, $w\sim\mathcal N(0,I_d)$, zero bias, inputs uniform on $S^{d-1}$, $t=0.3$, $\delta=0.01$. All numbers come from \texttt{calibration/kappa\_tanh.py} (fixed seeds).

\begin{table}[!htbp]
\centering
\caption{Estimated concentration-step width certificates \eqref{eq:pstar} for $\tanh$ at $t=0.3$, $\delta=0.01$, with $v_{\mathrm{std}}=4m^2$ versus the data-conditional row-covariance proxy $v=\|\Rowcov(X)\|_\op$ (min / median / max over ten input draws). $k^\star$ counts independent neuron pairs; the total width is $p^\star=2k^\star$. Numerical estimates, not certified.}
\label{tab:width}
\small
\begin{tabular}{@{}lcc@{}}
\toprule
 & $(m,d)=(16,8)$ & $(m,d)=(32,16)$ \\
\midrule
$\log(4m/\delta)$ & $\log 6400=8.7641$ & $\log 12800=9.4572$ \\
$k^\star_{\mathrm{std}}$, $v=4m^2$ & $8.7641\cdot(2048+6.4)/0.09=200{,}054$ & $9.4572\cdot(8192+12.8)/0.09=862{,}160$ \\
$p^\star_{\mathrm{std}}=2k^\star_{\mathrm{std}}$ & $400{,}108$ & $1{,}724{,}321$ \\
$v=\|\Rowcov(X)\|_\op$ & $16.05$ / $19.66$ / $25.62$ & $27.49$ / $33.56$ / $37.05$ \\
$\kappa=4m^2/v$ & $40.0$ / $52.1$ / $63.8$ & $110.6$ / $122.1$ / $149.0$ \\
$k^\star$, $v=\|\Rowcov(X)\|_\op$ & $3{,}749$ / $4{,}451$ / $5{,}612$ & $7{,}122$ / $8{,}398$ / $9{,}131$ \\
\quad median, by \eqref{eq:pstar} & $8.7641\cdot(2\cdot19.656+6.4)/0.09=4{,}451$ & $9.4572\cdot(2\cdot33.562+12.8)/0.09=8{,}398$ \\
$p^\star=2k^\star$ & $7{,}499$ / $8{,}903$ / $11{,}225$ & $14{,}244$ / $16{,}797$ / $18{,}263$ \\
$p^\star_{\mathrm{std}}/p^\star$ & $35.6$ / $45.0$ / $53.4$ & $94.4$ / $102.7$ / $121.1$ \\
\midrule
$v_+$ (heuristic margin) & $16.28$ / $19.88$ / $25.83$ & $28.34$ / $34.41$ / $37.90$ \\
$p^\star$ at $v_+$ (min--max) & $7{,}588$--$11{,}309$ & $14{,}603$--$18{,}618$ \\
$p^\star_{\mathrm{std}}/p^\star$ at $v_+$ & $35.4$ / $44.5$ / $52.7$ & $92.6$ / $100.5$ / $118.1$ \\
\bottomrule
\end{tabular}
\end{table}

\paragraph{Results.} At $(m,d)=(32,16)$ the estimated certificate drops from a total width of about $1.7\cdot10^6$ to $1.4\cdot10^4$--$1.8\cdot10^4$ neurons, $94$--$121\times$ smaller (median $103\times$); at $(16,8)$ it is $36$--$53\times$ smaller (median $45\times$). With the heuristic margin $v_+$ in place of $\hat v$ the savings are $93$--$118\times$ and $35$--$53\times$. In the variance-dominated regime the ratio approaches $\kappa$; the linear term $\frac43mt$ in \eqref{eq:pstar} makes it somewhat smaller at $t=0.3$. The deviation that coercivity actually needs is $t\approx\lambda_0/2$, which for $\tanh$ is much smaller than $0.3$; there, with each draw's own $\lambda_0(X)$, the ratio is $39.5$--$62.5\times$ (median $51.3\times$) at $(16,8)$ and $108.9$--$144.2\times$ (median $118.9\times$) at $(32,16)$, while both certificates become much larger (total widths: the standard one is $3.3\cdot10^7$--$6.1\cdot10^7$ at $(16,8)$ and $8.5\cdot10^7$--$2.1\cdot10^8$ at $(32,16)$; the row-covariance one $5.5\cdot10^5$--$1.4\cdot10^6$ and $5.9\cdot10^5$--$1.9\cdot10^6$; \texttt{kappa\_tanh\_output.txt}).

\paragraph{Remarks.} (i) Using the centred covariance $\E(A-\E A)^2$ instead of $\E A^2$ gives larger $\kappa$ (median $141.1$ and $213.8$); we report the conservative uncentred value. (ii) Averaging over the inputs as well (the ``population'' proxy) is misleading for $\tanh$: because $\tanh$ is odd and the input law is symmetric, the off-diagonal entries of $\E A^2$ vanish on average and the population $\kappa$ grows linearly in $m$ ($268.4$ at $(16,8)$, $651.9$ at $(32,16)$; \Cref{app:calibration}). For a fixed dataset these off-diagonal entries do not vanish, which is why we report the data-conditional values. (iii) The numbers quoted in the version submitted for review and during the review discussion ($\kappa\approx84$--$113$; certificate savings of $96\times$ and $67\times$; $\lambda_0\in[0.16,0.21]$ and $[0.19,0.23]$) were computed for clipped ReLU, which is not covered by the formal theorem; they are superseded by the $\tanh$ values in \Cref{tab:width,tab:assumptions}.

\section{Parameter norm and the linearisation error}\label{sec:norm}

Lazy-training analyses~\cite{chizat2018lazy} bound the gap between the network and its linearisation at $\theta_0$ by a Taylor remainder, $\frac12\sup\|D^2f\|\cdot\|\theta-\theta_0\|^2$, combined with a width-independent bound on $\|\theta-\theta_0\|$. Both factors depend on the norm on the parameter space, which informal proofs leave implicit.

\paragraph{Sup norm: an obstruction to the proof route (paper calculation, not in Lean).} Take the sup norm $\|(a,w)\|_\infty=\max(\max_j|a_j|,\max_j\|w_j\|_2)$. The mixed $a$--$w$ part of the second derivative of $f_\theta(x)$ is $\frac{2}{\sqrt p}\sum_j\delta a_j\,\sigma'(\langle w_j,x\rangle+b_j)\,\langle\delta w_j,x\rangle$. At $w_j=0$, $b_j=0$ with the perturbation $\delta a_j=1$, $\delta w_j=x/\|x\|$ for all $j$, every summand equals $\sigma'(0)\|x\|$, so the bilinear form equals $2\sqrt p\,\sigma'(0)\|x\|$ on the unit sup-norm ball; for $\tanh$, $\sigma'(0)=1$. A matching upper bound of order $\sqrt p$ follows from the triangle inequality, so the Hessian operator norm in this geometry is $\Theta(\sqrt p)$. Combined with a movement bound of order one, the sup-norm Taylor route gives a remainder of order $\sqrt p$ rather than $1/\sqrt p$. This shows only that this \emph{proof route} fails; it does not show that linearisation fails along gradient-flow trajectories.

The artifact contains a sup-norm-parameter version of the linearisation bound (\lean{ntk_network_approx_kernel_expansion_analytical}) whose displayed $O(1/\sqrt p)$ bound takes the segment Hessian bound $\|D^2f\|\le L_0/\sqrt p$ as a hypothesis (\lean{hHess}). By the calculation above, this hypothesis cannot be met with a width-independent $L_0$ near points where $\sigma'\neq0$, which is why we do not rely on that version.

\paragraph{$L^2$ geometry.} On the Euclidean ($L^2$) parameter space, Cauchy--Schwarz over the $p$ neurons gives $\|D^2f_\theta(x)\|\le L_0/\sqrt p$ with $L_0=C_{\sigma''}A\,C_x^2+2C_{\sigma'}C_x$, where $A$ bounds the output weights and $C_x\ge\|x\|$; this Hessian bound is proved in Lean with no Hessian hypothesis. Combined with the energy argument of~\cite[App.~B.3]{chizat2018lazy}, which bounds the movement directly in $L^2$ by $\|\theta(T)-\theta_0\|_2\le2\sqrt{K_{\max}}\,r_0/\rho$, it yields
\begin{equation}\label{eq:l2}
\bigl|f_{\theta(T)}(x)-f_{\theta_0}(x)-Df_{\theta_0}(x)[\theta(T)-\theta_0]\bigr|\ \le\ \frac{L_0}{2\sqrt p}\Bigl(\frac{2\sqrt{K_{\max}}\,r_0}{\rho}\Bigr)^2 = O\Bigl(\frac{1}{\sqrt p}\Bigr).
\end{equation}

\paragraph{Status of \eqref{eq:l2}.} The Lean theorem \lean{chizat_bach_analytical_L2_general} proves \eqref{eq:l2} for any $C^2$ activation with bounded $\sigma'$ and $\sigma''$ (for $\tanh$, $|\tanh'|,|\tanh''|\le1$ are proved). It is a \emph{conditional} statement about an abstract $L^2$ flow $\dot\theta=-G$: the residual decay $\|r(t)\|\le r_0e^{-\rho t/2}$, the gradient bound $\|G(t)\|\le\sqrt{K_{\max}}\|r(t)\|$ and a bound on the output weights along the segment $[\theta_0,\theta(T)]$ are hypotheses. They are of the kind that \Cref{thm:main} supplies (with $\rho=\lambda_0/2$ and $K_{\max}$ an upper bound on $\lmax$ of the kernel), but the two results are not yet composed in Lean, because \Cref{thm:main} is stated on the sup-norm parameter type. The older names \lean{chizat_bach_analytical_softplus_L2_unconditional} and \lean{chizat_bach_energy_movement_L2_bounded} denote the special case $\rho=K_{\max}$, which for an NTK flow requires $\lmax/\lmin\lesssim2$ of the kernel; ``unconditional'' in those names refers only to the Hessian bound.

\section{Related work}\label{sec:related}

\paragraph{Formalised learning theory.} AI4SLT~\cite{zhang2026ai4slt} formalises empirical-process theory, and \citet{sonoda2025rademacher} formalise Rademacher-complexity bounds with Dudley's entropy integral, both in Lean~4. TorchLean~\cite{torchlean2026} gives executable neural-network semantics in Lean. Earlier proof-assistant work on machine learning focused on program correctness and generalisation guarantees~\cite{selsam2017developing,bagnall2019certifying}; \citet{buzzard2020formalising} illustrate the reach of Lean for research mathematics. We are not aware of a previous machine-checked finite-width NTK convergence theorem.

\paragraph{NTK theory.} The NTK was introduced by \citet{jacot2018neural}; finite-width convergence analyses include~\cite{du2019gradient,du2019deep,allenzhu2019convergence,zou2020gradient,li2018learning,arora2019finegrained}, and \citet{chizat2018lazy} characterise lazy training and its limits. Related lines study exact infinite-width kernels~\cite{arora2019exact,lee2018deep,cho2009kernel}, inductive bias~\cite{bietti2019inductive}, mean-field limits~\cite{mei2018mean,sirignano2020mean} and the gap between linearised and trained networks~\cite{ghorbani2021linearized}; see~\cite{bach2024learning} for a textbook treatment.

\paragraph{Matrix concentration.} Matrix Bernstein inequalities go back to~\cite{ahlswede2002strong} and were refined in~\cite{oliveira2010sums,tropp2012user,minsker2017some}; the row-covariance proxy of \Cref{thm:tightened} is the standard variance statistic in~\cite[\S6.1]{tropp2015introduction}. Our contribution there is the verified, proxy-parametric carrier and the data-conditional quantification on NTK instances, not the inequality.

\section{Limitations}\label{sec:limitations}

\paragraph{Narrow practical impact.} The main limitation is practical reach. \Cref{thm:main} is a lazy-regime result: its drift condition requires widths of order $m^4/\lambda_0^4$, which for our small $\tanh$ instances is $10^{15}$--$10^{16}$ neurons (\Cref{tab:assumptions}), far beyond any network that is trained in practice, and lazy training is known not to describe the feature-learning behaviour of practical networks~\cite{chizat2018lazy,ghorbani2021linearized}. The width-certificate improvement of \Cref{sec:rowcov} concerns one step of the argument; it makes theory-driven width prescriptions for that step $36$--$121\times$ tighter on our instances (medians $45\times$ and $103\times$), but it does not bring the end-to-end theorem to practical widths. Consequently the results do not directly improve a data-mining pipeline; their value is as verified, reusable infrastructure and as an audit of the theory, and downstream use (for example in formal analyses of kernel-based models) remains to be demonstrated.

\paragraph{Scope of the formal results.} The architecture is a two-layer network with frozen biases, squared loss and paired initialisation. Non-smooth activations (ReLU, clipped ReLU) are not covered; extending the chain via $C^{1,1}$ or mollification arguments is future work. The data assumption $\lambda_0>0$ is not proved for any concrete initialisation law. The row-covariance proxy is not yet composed into \Cref{thm:main}, and the $L^2$ linearisation bound is not yet composed with it either. Graph, deep, attention-based and Bayesian NTK variants~\cite{du2019gntk,hron2020transformer,arora2019exact} are outside the scope of this paper.

\paragraph{Numerics.} The values of $\kappa$, the (estimated) width certificates and $\lambda_0$ are Monte Carlo and quadrature estimates on synthetic sphere-uniform inputs; they are not certified by Lean and they depend on the activation, the bias law and the data.

\section{Conclusion}\label{sec:conclusion}

We machine-checked the finite-width NTK lazy-training argument for a two-layer network, from matrix Bernstein concentration to existence, uniqueness and exponential convergence of gradient flow, with $\lambda_0>0$ as the only data assumption and $\tanh$ as a Lean-verified instance. Writing the argument in Lean made its variance proxy and its parameter norm explicit, which exposed a $36$--$121\times$ slack in standard concentration-step width certificates (numerically, for data-conditional $\tanh$ instances) and showed that the $O(1/\sqrt p)$ linearisation route depends on the $L^2$ geometry. At the same time, the formal theorem lives in a regime of astronomically large widths, so its direct practical impact is narrow: it certifies the classical theory rather than extending it to realistic networks. Composing the sharper proxy and the $L^2$ linearisation bound into the end-to-end theorem, covering non-smooth activations, and connecting the verified components to concrete downstream analyses are the next steps.

\paragraph{Acknowledgements and disclosure.} The Lean development was produced with an AI-assisted workflow under human supervision; every theorem is checked by the Lean kernel.

\bibliographystyle{plainnat}
\bibliography{references}

\begin{thebibliography}{32}
\providecommand{\natexlab}[1]{#1}
\providecommand{\url}[1]{\texttt{#1}}
\expandafter\ifx\csname urlstyle\endcsname\relax
  \providecommand{\doi}[1]{doi: #1}\else
  \providecommand{\doi}{doi: \begingroup \urlstyle{rm}\Url}\fi

\bibitem[Ahlswede and Winter(2002)]{ahlswede2002strong}
Rudolf Ahlswede and Andreas Winter.
\newblock Strong converse for identification via quantum channels.
\newblock \emph{IEEE Transactions on Information Theory}, 48\penalty0 (3):\penalty0 569--579, 2002.

\bibitem[Allen-Zhu et~al.(2019)Allen-Zhu, Li, and Song]{allenzhu2019convergence}
Zeyuan Allen-Zhu, Yuanzhi Li, and Zhao Song.
\newblock A convergence theory for deep learning via over-parameterization.
\newblock In \emph{International Conference on Machine Learning (ICML)}, 2019.

\bibitem[Arora et~al.(2019{\natexlab{a}})Arora, Du, Hu, Li, Salakhutdinov, and Wang]{arora2019exact}
Sanjeev Arora, Simon~S. Du, Wei Hu, Zhiyuan Li, Ruslan Salakhutdinov, and Ruosong Wang.
\newblock On exact computation with an infinitely wide neural net.
\newblock In \emph{Advances in Neural Information Processing Systems (NeurIPS)}, 2019{\natexlab{a}}.

\bibitem[Arora et~al.(2019{\natexlab{b}})Arora, Du, Hu, Li, and Wang]{arora2019finegrained}
Sanjeev Arora, Simon~S. Du, Wei Hu, Zhiyuan Li, and Ruosong Wang.
\newblock Fine-grained analysis of optimization and generalization for overparameterized two-layer neural networks.
\newblock In \emph{Proceedings of the 36th International Conference on Machine Learning (ICML)}, volume~97 of \emph{Proceedings of Machine Learning Research}, pages 322--332, 2019{\natexlab{b}}.

\bibitem[Bach(2024)]{bach2024learning}
Francis Bach.
\newblock \emph{Learning Theory from First Principles}.
\newblock MIT Press, 2024.

\bibitem[Bagnall and Stewart(2019)]{bagnall2019certifying}
Alexander Bagnall and Gordon Stewart.
\newblock Certifying the true error: Machine learning in {C}oq with verified generalization guarantees.
\newblock In \emph{AAAI Conference on Artificial Intelligence}, 2019.

\bibitem[Bietti and Mairal(2019)]{bietti2019inductive}
Alberto Bietti and Julien Mairal.
\newblock On the inductive bias of neural tangent kernels.
\newblock In \emph{Advances in Neural Information Processing Systems (NeurIPS)}, 2019.

\bibitem[Buzzard et~al.(2020)Buzzard, Commelin, and Massot]{buzzard2020formalising}
Kevin Buzzard, Johan Commelin, and Patrick Massot.
\newblock Formalising perfectoid spaces.
\newblock In \emph{International Conference on Certified Programs and Proofs (CPP)}, 2020.

\bibitem[Chizat et~al.(2019)Chizat, Oyallon, and Bach]{chizat2018lazy}
L\'ena\"ic Chizat, Edouard Oyallon, and Francis Bach.
\newblock On lazy training in differentiable programming.
\newblock In \emph{Advances in Neural Information Processing Systems (NeurIPS)}, 2019.

\bibitem[Cho and Saul(2009)]{cho2009kernel}
Youngmin Cho and Lawrence~K. Saul.
\newblock Kernel methods for deep learning.
\newblock In \emph{Advances in Neural Information Processing Systems}, 2009.

\bibitem[de~Moura and Ullrich(2021)]{demoura2021lean4}
Leonardo de~Moura and Sebastian Ullrich.
\newblock The {L}ean 4 theorem prover and programming language.
\newblock In \emph{Automated Deduction --- CADE 28}, pages 625--635, 2021.

\bibitem[Du et~al.(2019{\natexlab{a}})Du, Hou, Salakhutdinov, Poczos, Wang, and Xu]{du2019gntk}
Simon~S. Du, Kangcheng Hou, Ruslan Salakhutdinov, Barnabas Poczos, Ruosong Wang, and Keyulu Xu.
\newblock Graph neural tangent kernel: Fusing graph neural networks with graph kernels.
\newblock In \emph{Advances in Neural Information Processing Systems (NeurIPS)}, 2019{\natexlab{a}}.

\bibitem[Du et~al.(2019{\natexlab{b}})Du, Lee, Li, Wang, and Zhai]{du2019deep}
Simon~S. Du, Jason~D. Lee, Haochuan Li, Liwei Wang, and Xiyu Zhai.
\newblock Gradient descent finds global minima of deep neural networks.
\newblock In \emph{Proceedings of the 36th International Conference on Machine Learning (ICML)}, volume~97 of \emph{Proceedings of Machine Learning Research}, pages 1675--1685, 2019{\natexlab{b}}.

\bibitem[Du et~al.(2019{\natexlab{c}})Du, Zhai, P\'oczos, and Singh]{du2019gradient}
Simon~S. Du, Xiyu Zhai, Barnab\'as P\'oczos, and Aarti Singh.
\newblock Gradient descent provably optimizes over-parameterized neural networks.
\newblock In \emph{International Conference on Learning Representations (ICLR)}, 2019{\natexlab{c}}.

\bibitem[George et~al.(2026)George, Cruden, Zhong, Zhang, and Anandkumar]{torchlean2026}
Robert~Joseph George, Jennifer Cruden, Xiangru Zhong, Huan Zhang, and Anima Anandkumar.
\newblock {T}orch{L}ean: Formalizing neural networks in {L}ean, 2026.
\newblock arXiv:2602.22631; poster at the ICML 2026 AI for Math Workshop.

\bibitem[Ghorbani et~al.(2021)Ghorbani, Mei, Misiakiewicz, and Montanari]{ghorbani2021linearized}
Behrooz Ghorbani, Song Mei, Theodor Misiakiewicz, and Andrea Montanari.
\newblock Linearized two-layers neural networks in high dimension.
\newblock \emph{Annals of Statistics}, 49\penalty0 (2):\penalty0 1029--1054, 2021.

\bibitem[Hron et~al.(2020)Hron, Bahri, Sohl-Dickstein, and Novak]{hron2020transformer}
Jiri Hron, Yasaman Bahri, Jascha Sohl-Dickstein, and Roman Novak.
\newblock Infinite attention: {NNGP} and {NTK} for deep attention networks.
\newblock In \emph{International Conference on Machine Learning (ICML)}, 2020.

\bibitem[Jacot et~al.(2018)Jacot, Gabriel, and Hongler]{jacot2018neural}
Arthur Jacot, Franck Gabriel, and Cl\'ement Hongler.
\newblock Neural tangent kernel: Convergence and generalization in neural networks.
\newblock In \emph{Advances in Neural Information Processing Systems (NeurIPS)}, 2018.

\bibitem[Lee et~al.(2018)Lee, Bahri, Novak, Schoenholz, Pennington, and Sohl-Dickstein]{lee2018deep}
Jaehoon Lee, Yasaman Bahri, Roman Novak, Samuel~S. Schoenholz, Jeffrey Pennington, and Jascha Sohl-Dickstein.
\newblock Deep neural networks as gaussian processes.
\newblock In \emph{International Conference on Learning Representations}, 2018.

\bibitem[Li and Liang(2018)]{li2018learning}
Yuanzhi Li and Yingyu Liang.
\newblock Learning overparameterized neural networks via stochastic gradient descent on structured data.
\newblock In \emph{Advances in Neural Information Processing Systems}, 2018.

\bibitem[Lieb(1973)]{lieb1973convex}
Elliott~H. Lieb.
\newblock Convex trace functions and the {W}igner-{Y}anase-{D}yson conjecture.
\newblock \emph{Advances in Mathematics}, 11\penalty0 (3):\penalty0 267--288, 1973.

\bibitem[Mei et~al.(2018)Mei, Montanari, and Nguyen]{mei2018mean}
Song Mei, Andrea Montanari, and Phan-Minh Nguyen.
\newblock A mean field view of the landscape of two-layer neural networks.
\newblock \emph{Proceedings of the National Academy of Sciences}, 115\penalty0 (33):\penalty0 E7665--E7671, 2018.

\bibitem[Minsker(2017)]{minsker2017some}
Stanislav Minsker.
\newblock On some extensions of {B}ernstein's inequality for self-adjoint operators.
\newblock \emph{Statistics and Probability Letters}, 127:\penalty0 111--119, 2017.

\bibitem[Oliveira(2010)]{oliveira2010sums}
Roberto~Imbuzeiro Oliveira.
\newblock Sums of random {H}ermitian matrices and an inequality by {R}udelson.
\newblock \emph{Electronic Communications in Probability}, 15:\penalty0 203--212, 2010.

\bibitem[Selsam et~al.(2017)Selsam, Liang, and Dill]{selsam2017developing}
Daniel Selsam, Percy Liang, and David~L. Dill.
\newblock Developing bug-free machine learning systems with formal mathematics.
\newblock In \emph{International Conference on Machine Learning}, 2017.

\bibitem[Sirignano and Spiliopoulos(2020)]{sirignano2020mean}
Justin Sirignano and Konstantinos Spiliopoulos.
\newblock Mean field analysis of neural networks: A central limit theorem.
\newblock \emph{Stochastic Processes and their Applications}, 130\penalty0 (3):\penalty0 1820--1852, 2020.

\bibitem[Sonoda et~al.(2026)Sonoda, Kasaura, Mizuno, Tsukamoto, and Onda]{sonoda2025rademacher}
Sho Sonoda, Kazumi Kasaura, Yuma Mizuno, Kei Tsukamoto, and Naoto Onda.
\newblock {L}ean formalization of generalization error bound by {R}ademacher complexity and {D}udley's entropy integral.
\newblock In \emph{17th International Conference on Interactive Theorem Proving (ITP 2026)}, volume 382 of \emph{Leibniz International Proceedings in Informatics (LIPIcs)}, pages 8:1--8:17, 2026.
\newblock \doi{10.4230/LIPIcs.ITP.2026.8}.

\bibitem[{The mathlib Community}(2020)]{mathlib2024}
{The mathlib Community}.
\newblock The {L}ean mathematical library.
\newblock In \emph{Proceedings of the 9th ACM SIGPLAN International Conference on Certified Programs and Proofs (CPP)}, pages 367--381, 2020.
\newblock \doi{10.1145/3372885.3373824}.

\bibitem[Tropp(2012)]{tropp2012user}
Joel~A. Tropp.
\newblock User-friendly tail bounds for sums of random matrices.
\newblock \emph{Foundations of Computational Mathematics}, 12\penalty0 (4):\penalty0 389--434, 2012.

\bibitem[Tropp(2015)]{tropp2015introduction}
Joel~A. Tropp.
\newblock An introduction to matrix concentration inequalities.
\newblock \emph{Foundations and Trends in Machine Learning}, 8\penalty0 (1--2):\penalty0 1--230, 2015.

\bibitem[Zhang et~al.(2026)Zhang, Lee, and Liu]{zhang2026ai4slt}
Yuanhe Zhang, Jason~D. Lee, and Fanghui Liu.
\newblock {AI4SLT}: Empirical processes in {L}ean 4 for formal statistical learning theory.
\newblock In \emph{Proceedings of the 43rd International Conference on Machine Learning (ICML)}, 2026.

\bibitem[Zou et~al.(2020)Zou, Cao, Zhou, and Gu]{zou2020gradient}
Difan Zou, Yuan Cao, Dongruo Zhou, and Quanquan Gu.
\newblock Gradient descent optimizes over-parameterized deep {ReLU} networks.
\newblock \emph{Machine Learning}, 109\penalty0 (3):\penalty0 467--492, 2020.

\end{thebibliography}

\appendix

\section{Lean statements}\label{app:listings}

The listings below reproduce the Lean declarations (statements only, without proofs) of the main results.

\subsection{End-to-end theorem}
\noindent\lean{ProbabilityTheory.fullNet_gradient_flow_lazy_training_whp_exists_uniform}
\begin{leanlisting}
theorem fullNet_gradient_flow_lazy_training_whp_exists_uniform
    {n k : ℕ} [Nonempty (Fin n)] (hk : 0 < k)
    (σ : ℝ → ℝ) (hσ : Differentiable ℝ σ)
    {M M' : ℝ} (hM : 0 < M) (hM' : 0 < M')
    (hσ_bdd : ∀ z, |σ z| ≤ M) (hσ'_bdd : ∀ z, |deriv σ z| ≤ M')
    {Lσ' : NNReal} (hσ'_lip : LipschitzWith Lσ' (deriv σ))
    (y : Fin n → ℝ) (xs : Fin n → EuclideanSpace ℝ (Fin d))
    {X : ℝ} (hX_pos : 0 < X) (hX : ∀ i, ‖xs i‖ ≤ X)
    {ν : Measure (EuclideanSpace ℝ (Fin d) × ℝ)} [IsProbabilityMeasure ν]
    {lam₀ : ℝ} (hlam : 0 < lam₀)
    (hpop : lam₀ • (1 : Matrix (Fin n) (Fin n) ℝ) ≤ populationFullNTK σ (deriv σ) xs ν)
    {δ : ℝ} (hδ : 0 < δ)
    (hWσ : ntkBernsteinWidth n (M ^ 2) (lam₀ / 4) δ ≤ k)
    (hWσ' : ntkBernsteinWidth n (M' ^ 2 * X ^ 2) (lam₀ / 4) δ ≤ k)
    (hW₁ : gfMovementBound n (k + k) (lam₀ / 2) M M' X
      ‖(EuclideanSpace.equiv (Fin n) ℝ).symm y‖ < 1)
    (hW₂ : (n : ℝ) * gfDriftConst M M' Lσ' X *
      gfMovementBound n (k + k) (lam₀ / 2) M M' X
        ‖(EuclideanSpace.equiv (Fin n) ℝ).symm y‖ < lam₀ / 4) :
    ∃ G : Set (Fin k → EuclideanSpace ℝ (Fin d) × ℝ), MeasurableSet G ∧
      1 - δ ≤ (Measure.pi (fun _ : Fin k => ν)).real G ∧
      ∀ ω ∈ G, ∀ T : ℝ,
        (∃ θ : ℝ → Param d (k + k), θ 0 = antisymInitParam ω ∧ ContinuousOn θ (Icc 0 T) ∧
          ∀ t ∈ Ico 0 T, HasDerivWithinAt θ
            (gfFieldFullNet σ (antisymInitBias ω) y xs (θ t)) (Ici t) t) ∧
        (∀ θ θ' : ℝ → Param d (k + k), θ 0 = antisymInitParam ω →
          θ' 0 = antisymInitParam ω → ContinuousOn θ (Icc 0 T) →
          ContinuousOn θ' (Icc 0 T) →
          (∀ t ∈ Ico 0 T, HasDerivWithinAt θ
            (gfFieldFullNet σ (antisymInitBias ω) y xs (θ t)) (Ici t) t) →
          (∀ t ∈ Ico 0 T, HasDerivWithinAt θ'
            (gfFieldFullNet σ (antisymInitBias ω) y xs (θ' t)) (Ici t) t) →
          EqOn θ θ' (Icc 0 T)) ∧
        (∀ θ : ℝ → Param d (k + k), θ 0 = antisymInitParam ω → ContinuousOn θ (Icc 0 T) →
          (∀ t ∈ Ico 0 T, HasDerivWithinAt θ
            (gfFieldFullNet σ (antisymInitBias ω) y xs (θ t)) (Ici t) t) →
          ∀ t ∈ Icc 0 T,
            ‖residualFullNet σ (antisymInitBias ω) y xs (θ t)‖ ^ 2 ≤
              ‖(EuclideanSpace.equiv (Fin n) ℝ).symm y‖ ^ 2 *
                Real.exp (-(lam₀ / 2 * t)))
\end{leanlisting}

\noindent The width conditions use the following definitions.
\begin{leanlisting}
noncomputable def ntkBernsteinWidth (n : ℕ) (A s δ : ℝ) : ℝ :=
  2 * (4 * (n : ℝ) ^ 2 * A ^ 2 + 2 * (n : ℝ) * A * s / 3) / s ^ 2 *
    Real.log (8 * (n : ℝ) / δ)
\end{leanlisting}
\begin{leanlisting}
noncomputable def gfMovementBound (n m : ℕ) (ρ M M' X R₀ : ℝ) : ℝ :=
  2 / ρ * (1 / Real.sqrt m * (M + 2 * M' * X) * Real.sqrt n * R₀)
\end{leanlisting}
\begin{leanlisting}
noncomputable def gfDriftConst (M M' Lσ' X : ℝ) : ℝ :=
  2 * M * M' * X + 2 * 2 * M' ^ 2 * X ^ 2 + 2 * 2 ^ 2 * M' * Lσ' * X * X ^ 2
\end{leanlisting}

\subsection{The \texorpdfstring{$\tanh$}{tanh} instance}
\noindent\lean{NTK.Tanh.hasDerivAt_tanh}, \lean{NTK.Tanh.lipschitzWith_deriv_tanh},\\ \lean{NTK.Tanh.abs_iteratedDeriv_two_tanh_le_one}
\begin{leanlisting}
theorem hasDerivAt_tanh (x : ℝ) : HasDerivAt Real.tanh (1 - Real.tanh x ^ 2) x
\end{leanlisting}
\begin{leanlisting}
theorem lipschitzWith_deriv_tanh : LipschitzWith 1 (deriv Real.tanh)
\end{leanlisting}
\begin{leanlisting}
theorem abs_iteratedDeriv_two_tanh_le_one (x : ℝ) : |iteratedDeriv 2 Real.tanh x| ≤ 1
\end{leanlisting}
\noindent\lean{ProbabilityTheory.fullNet_gradient_flow_lazy_training_whp_exists_uniform_tanh}
\begin{leanlisting}
theorem fullNet_gradient_flow_lazy_training_whp_exists_uniform_tanh
    {n k : ℕ} [Nonempty (Fin n)] (hk : 0 < k)
    (y : Fin n → ℝ) (xs : Fin n → EuclideanSpace ℝ (Fin d))
    {X : ℝ} (hX_pos : 0 < X) (hX : ∀ i, ‖xs i‖ ≤ X)
    {ν : Measure (EuclideanSpace ℝ (Fin d) × ℝ)} [IsProbabilityMeasure ν]
    {lam₀ : ℝ} (hlam : 0 < lam₀)
    (hpop : lam₀ • (1 : Matrix (Fin n) (Fin n) ℝ) ≤
      populationFullNTK Real.tanh (deriv Real.tanh) xs ν)
    {δ : ℝ} (hδ : 0 < δ)
    (hWσ : ntkBernsteinWidth n (1 ^ 2) (lam₀ / 4) δ ≤ k)
    (hWσ' : ntkBernsteinWidth n (1 ^ 2 * X ^ 2) (lam₀ / 4) δ ≤ k)
    (hW₁ : gfMovementBound n (k + k) (lam₀ / 2) 1 1 X
      ‖(EuclideanSpace.equiv (Fin n) ℝ).symm y‖ < 1)
    (hW₂ : (n : ℝ) * gfDriftConst 1 1 (1 : NNReal) X *
      gfMovementBound n (k + k) (lam₀ / 2) 1 1 X
        ‖(EuclideanSpace.equiv (Fin n) ℝ).symm y‖ < lam₀ / 4) :
    ∃ G : Set (Fin k → EuclideanSpace ℝ (Fin d) × ℝ), MeasurableSet G ∧
      1 - δ ≤ (Measure.pi (fun _ : Fin k => ν)).real G ∧
      ∀ ω ∈ G, ∀ T : ℝ,
        (∃ θ : ℝ → Param d (k + k), θ 0 = antisymInitParam ω ∧ ContinuousOn θ (Icc 0 T) ∧
          ∀ t ∈ Ico 0 T, HasDerivWithinAt θ
            (gfFieldFullNet Real.tanh (antisymInitBias ω) y xs (θ t)) (Ici t) t) ∧
        (∀ θ θ' : ℝ → Param d (k + k), θ 0 = antisymInitParam ω →
          θ' 0 = antisymInitParam ω → ContinuousOn θ (Icc 0 T) →
          ContinuousOn θ' (Icc 0 T) →
          (∀ t ∈ Ico 0 T, HasDerivWithinAt θ
            (gfFieldFullNet Real.tanh (antisymInitBias ω) y xs (θ t)) (Ici t) t) →
          (∀ t ∈ Ico 0 T, HasDerivWithinAt θ'
            (gfFieldFullNet Real.tanh (antisymInitBias ω) y xs (θ' t)) (Ici t) t) →
          EqOn θ θ' (Icc 0 T)) ∧
        (∀ θ : ℝ → Param d (k + k), θ 0 = antisymInitParam ω → ContinuousOn θ (Icc 0 T) →
          (∀ t ∈ Ico 0 T, HasDerivWithinAt θ
            (gfFieldFullNet Real.tanh (antisymInitBias ω) y xs (θ t)) (Ici t) t) →
          ∀ t ∈ Icc 0 T,
            ‖residualFullNet Real.tanh (antisymInitBias ω) y xs (θ t)‖ ^ 2 ≤
              ‖(EuclideanSpace.equiv (Fin n) ℝ).symm y‖ ^ 2 *
                Real.exp (-(lam₀ / 2 * t)))
\end{leanlisting}

\subsection{Components}
\noindent\lean{ProbabilityTheory.fullNet_gradient_flow_lazy_training} (bootstrap)
\begin{leanlisting}
theorem fullNet_gradient_flow_lazy_training
    [Nonempty (Fin n)] (hm : 0 < m)
    (σ : ℝ → ℝ) (hσ : Differentiable ℝ σ)
    {M M' : ℝ} (hσ_bdd : ∀ z, |σ z| ≤ M) (hσ'_bdd : ∀ z, |deriv σ z| ≤ M')
    {Lσ' : NNReal} (hσ'_lip : LipschitzWith Lσ' (deriv σ))
    (b : Fin m → ℝ) (y : Fin n → ℝ) (xs : Fin n → EuclideanSpace ℝ (Fin d))
    {X : ℝ} (hX : ∀ i, ‖xs i‖ ≤ X)
    (θ : ℝ → Param d m) {T : ℝ}
    (hθ_cont : ContinuousOn θ (Icc 0 T))
    (hθ_der : ∀ t ∈ Ico 0 T,
      HasDerivWithinAt θ (gfFieldFullNet σ b y xs (θ t)) (Ici t) t)
    (ha₀ : ∀ j, |(θ 0).1 j| ≤ 1)
    {ρ : ℝ} (hρ : 0 < ρ)
    (hK₀ : ρ • (1 : Matrix (Fin n) (Fin n) ℝ) ≤
      fullTrainingKernel σ (deriv σ) b (θ 0) xs)
    {R₀ : ℝ} (hR₀ : ‖residualFullNet σ b y xs (θ 0)‖ ≤ R₀)
    (hW₁ : gfMovementBound n m ρ M M' X R₀ < 1)
    (hW₂ : (n : ℝ) * gfDriftConst M M' Lσ' X * gfMovementBound n m ρ M M' X R₀ < ρ / 2) :
    ∀ t ∈ Icc 0 T,
      ‖residualFullNet σ b y xs (θ t)‖ ^ 2 ≤
          ‖residualFullNet σ b y xs (θ 0)‖ ^ 2 * Real.exp (-(ρ * t)) ∧
        ∀ j, dist ((θ t).1 j, (θ t).2 j) ((θ 0).1 j, (θ 0).2 j) ≤
          gfMovementBound n m ρ M M' X R₀
\end{leanlisting}
\noindent\lean{ProbabilityTheory.fullTrainingKernel_antisymInit_coercive_whp} (initial coercivity)
\begin{leanlisting}
theorem fullTrainingKernel_antisymInit_coercive_whp {n k : ℕ} [Nonempty (Fin n)]
    (hk : 0 < k) {σ σ' : ℝ → ℝ}
    (hσ_cont : Continuous σ) {M : ℝ} (hM : 0 < M) (hσ_bdd : ∀ z, |σ z| ≤ M)
    (hσ'_cont : Continuous σ') {M' : ℝ} (hM' : 0 < M') (hσ'_bdd : ∀ z, |σ' z| ≤ M')
    (xs : Fin n → EuclideanSpace ℝ (Fin d)) {X : ℝ} (hX_pos : 0 < X)
    (hX : ∀ i, ‖xs i‖ ≤ X)
    {ν : Measure (EuclideanSpace ℝ (Fin d) × ℝ)} [IsProbabilityMeasure ν]
    {lam₀ : ℝ} (hlam : 0 < lam₀)
    (hpop : lam₀ • (1 : Matrix (Fin n) (Fin n) ℝ) ≤ populationFullNTK σ σ' xs ν)
    {δ : ℝ} (hδ : 0 < δ)
    (hWσ : ntkBernsteinWidth n (M ^ 2) (lam₀ / 4) δ ≤ k)
    (hWσ' : ntkBernsteinWidth n (M' ^ 2 * X ^ 2) (lam₀ / 4) δ ≤ k) :
    1 - δ ≤ (Measure.pi (fun _ : Fin k => ν)).real
      {ω | (lam₀ / 2) • (1 : Matrix (Fin n) (Fin n) ℝ) ≤
        fullTrainingKernel σ σ' (antisymInitBias ω) (antisymInitParam ω) xs}
\end{leanlisting}
\noindent\lean{ProbabilityTheory.fderiv_sqLossFullNet_apply} and\\ \lean{ProbabilityTheory.residualFullNet_ode_on_Icc}
\begin{leanlisting}
theorem fderiv_sqLossFullNet_apply
    (σ : ℝ → ℝ) (hσ : Differentiable ℝ σ) (b : Fin m → ℝ) (y : Fin n → ℝ)
    (xs : Fin n → EuclideanSpace ℝ (Fin d)) (θ v : Param d m) :
    fderiv ℝ (sqLossFullNet σ b y xs) θ v =
      -paramInner (gfFieldFullNet σ b y xs θ) v
\end{leanlisting}
\begin{leanlisting}
theorem residualFullNet_ode_on_Icc
    (σ : ℝ → ℝ) (hσ : Differentiable ℝ σ) (b : Fin m → ℝ) (y : Fin n → ℝ)
    (xs : Fin n → EuclideanSpace ℝ (Fin d)) (hm : 0 < m)
    (θ : ℝ → Param d m) (T : ℝ)
    (hθ : ∀ t ∈ Set.Ico (0 : ℝ) T,
      HasDerivWithinAt θ (gfFieldFullNet σ b y xs (θ t)) (Set.Ici t) t) :
    ∀ t ∈ Set.Ico (0 : ℝ) T,
      HasDerivWithinAt (fun u => residualFullNet σ b y xs (θ u))
        (-(Matrix.toEuclideanLin (fullTrainingKernel σ (deriv σ) b (θ t) xs))
          (residualFullNet σ b y xs (θ t))) (Set.Ici t) t
\end{leanlisting}
\noindent\lean{ProbabilityTheory.exists_fullNet_gradient_flow} and\\ \lean{ProbabilityTheory.fullNet_gradient_flow_unique}
\begin{leanlisting}
theorem exists_fullNet_gradient_flow
    [Nonempty (Fin n)] (hm : 0 < m)
    (σ : ℝ → ℝ) (hσ : Differentiable ℝ σ)
    {M M' : ℝ} (hσ_bdd : ∀ z, |σ z| ≤ M) (hσ'_bdd : ∀ z, |deriv σ z| ≤ M')
    {Lσ' : NNReal} (hσ'_lip : LipschitzWith Lσ' (deriv σ))
    (b : Fin m → ℝ) (y : Fin n → ℝ) (xs : Fin n → EuclideanSpace ℝ (Fin d))
    {X : ℝ} (hX : ∀ i, ‖xs i‖ ≤ X)
    (θ₀ : Param d m) (ha₀ : ∀ j, |θ₀.1 j| ≤ 1)
    {ρ : ℝ} (hρ : 0 < ρ)
    (hK₀ : ρ • (1 : Matrix (Fin n) (Fin n) ℝ) ≤ fullTrainingKernel σ (deriv σ) b θ₀ xs)
    {R₀ : ℝ} (hR₀ : ‖residualFullNet σ b y xs θ₀‖ ≤ R₀)
    (hW₁ : gfMovementBound n m ρ M M' X R₀ < 1)
    (hW₂ : (n : ℝ) * gfDriftConst M M' Lσ' X * gfMovementBound n m ρ M M' X R₀ < ρ / 2)
    (T : ℝ) :
    ∃ θ : ℝ → Param d m, θ 0 = θ₀ ∧ ContinuousOn θ (Icc 0 T) ∧
      (∀ t ∈ Icc 0 T, HasDerivWithinAt θ (gfFieldFullNet σ b y xs (θ t)) (Icc 0 T) t) ∧
      (∀ t ∈ Ico 0 T, HasDerivWithinAt θ (gfFieldFullNet σ b y xs (θ t)) (Ici t) t) ∧
      ∀ t ∈ Icc 0 T, ‖θ t - θ₀‖ < 1
\end{leanlisting}
\begin{leanlisting}
theorem fullNet_gradient_flow_unique
    [Nonempty (Fin n)] (hm : 0 < m)
    (σ : ℝ → ℝ) (hσ : Differentiable ℝ σ)
    {M M' : ℝ} (hσ_bdd : ∀ z, |σ z| ≤ M) (hσ'_bdd : ∀ z, |deriv σ z| ≤ M')
    {Lσ' : NNReal} (hσ'_lip : LipschitzWith Lσ' (deriv σ))
    (b : Fin m → ℝ) (y : Fin n → ℝ) (xs : Fin n → EuclideanSpace ℝ (Fin d))
    {X : ℝ} (hX : ∀ i, ‖xs i‖ ≤ X)
    (θ₀ : Param d m) (ha₀ : ∀ j, |θ₀.1 j| ≤ 1)
    {ρ : ℝ} (hρ : 0 < ρ)
    (hK₀ : ρ • (1 : Matrix (Fin n) (Fin n) ℝ) ≤ fullTrainingKernel σ (deriv σ) b θ₀ xs)
    {R₀ : ℝ} (hR₀ : ‖residualFullNet σ b y xs θ₀‖ ≤ R₀)
    (hW₁ : gfMovementBound n m ρ M M' X R₀ < 1)
    (hW₂ : (n : ℝ) * gfDriftConst M M' Lσ' X * gfMovementBound n m ρ M M' X R₀ < ρ / 2)
    {T : ℝ} (θ θ' : ℝ → Param d m) (h0 : θ 0 = θ₀) (h0' : θ' 0 = θ₀)
    (hc : ContinuousOn θ (Icc 0 T)) (hc' : ContinuousOn θ' (Icc 0 T))
    (hd : ∀ t ∈ Ico 0 T, HasDerivWithinAt θ (gfFieldFullNet σ b y xs (θ t)) (Ici t) t)
    (hd' : ∀ t ∈ Ico 0 T, HasDerivWithinAt θ' (gfFieldFullNet σ b y xs (θ' t)) (Ici t) t) :
    EqOn θ θ' (Icc 0 T)
\end{leanlisting}

\subsection{Matrix Bernstein and the row-covariance proxy}
\noindent\lean{Matrix.lieb_tropp_concave}
\begin{leanlisting}
theorem lieb_tropp_concave
    {n : Type*} [Fintype n] [DecidableEq n] [Nonempty n]
    (H : Matrix n n ℂ) (hH : H.IsHermitian) :
    ConcaveOn ℝ
      {A : Matrix n n ℂ | IsStrictlyPositive A}
      (fun A => (Matrix.trace (NormedSpace.exp (H + CFC.log A) :
        Matrix n n ℂ)).re)
\end{leanlisting}
\noindent\lean{Matrix.bernstein_full}
\begin{leanlisting}
theorem bernstein_full
    {dn m : Type*} [Fintype dn] [DecidableEq dn] [Nonempty dn]
    [Fintype m] [DecidableEq m]
    {Ω : m → Type*} [∀ i, MeasurableSpace (Ω i)]
    (μ : ∀ i, MeasureTheory.Measure (Ω i))
    [∀ i, MeasureTheory.IsProbabilityMeasure (μ i)]
    (X : ∀ i, Ω i → Matrix dn dn ℂ)
    (hX : ∀ i ω, (X i ω).IsHermitian)
    (hmeas : ∀ i, MeasureTheory.AEStronglyMeasurable (X i) (μ i))
    (R : ℝ) (hR : 0 ≤ R) (hbound : ∀ i ω, ‖X i ω‖ ≤ R)
    (hcenter : ∀ i, ∫ x, X i x ∂μ i = 0)
    (σ2 : ℝ) (hσ2 : 0 < σ2)
    (hvar : ∑ i, ∫ x, X i x * X i x ∂μ i ≤ σ2 • (1 : Matrix dn dn ℂ))
    (t : ℝ) (ht : 0 < t)
    (hSum : ∀ ω : ∀ i, Ω i, (∑ i, X i (ω i)).IsHermitian)
    (hLamMeas : AEMeasurable
      (fun ω => Finset.sup' Finset.univ Finset.univ_nonempty (hSum ω).eigenvalues)
      (MeasureTheory.Measure.pi μ))
    (htrInt : MeasureTheory.Integrable
      (fun ω => (Matrix.trace (NormedSpace.exp
        (NTK.matrix_bernstein_theta t σ2 R • (∑ i, X i (ω i))))).re)
      (MeasureTheory.Measure.pi μ)) :
    (MeasureTheory.Measure.pi μ).real
      {ω | t ≤ Finset.sup' Finset.univ Finset.univ_nonempty (hSum ω).eigenvalues}
    ≤ NTK.matrix_bernstein_bound (Fintype.card dn) t σ2 R
\end{leanlisting}
\noindent\lean{Matrix.row_covariance_loewner_bound}
\begin{leanlisting}
theorem row_covariance_loewner_bound
    [Nonempty (Fin n)]
    {ι : Type*} [Fintype ι]
    (X : ι → Matrix (Fin n) (Fin n) ℂ)
    (R : ℝ) (hR : ∀ i, ‖X i‖ ≤ R) :
    (∑ i, X i * (X i)ᴴ)
      ≤ (R ^ 2 * (Fintype.card ι : ℝ)) •
          (1 : Matrix (Fin n) (Fin n) ℂ)
\end{leanlisting}
\noindent\lean{ProbabilityTheory.ntk_matrix_bernstein_tightened}
\begin{leanlisting}
theorem ntk_matrix_bernstein_tightened
    {dn m : Type*} [Fintype dn] [DecidableEq dn] [Nonempty dn]
    [Fintype m] [DecidableEq m]
    {Ω : m → Type*} [∀ i, MeasurableSpace (Ω i)]
    (μ : ∀ i, MeasureTheory.Measure (Ω i))
    [∀ i, MeasureTheory.IsProbabilityMeasure (μ i)]
    (X : ∀ i, Ω i → Matrix dn dn ℂ)
    (hX : ∀ i ω, (X i ω).IsHermitian)
    (hmeas : ∀ i, MeasureTheory.AEStronglyMeasurable (X i) (μ i))
    (hXX_int : ∀ i, MeasureTheory.Integrable
      (fun ω => X i ω * (X i ω)ᴴ) (μ i))
    (R : ℝ) (hR : 0 ≤ R) (hbound : ∀ i ω, ‖X i ω‖ ≤ R)
    (hcenter : ∀ i, ∫ x, X i x ∂μ i = 0)
    (hRowCov_ne :
      (∑ i, ∫ x, X i x * (X i x)ᴴ ∂μ i) ≠
        (0 : Matrix dn dn ℂ))
    (t : ℝ) (ht : 0 < t)
    (hSum : ∀ ω : ∀ i, Ω i, (∑ i, X i (ω i)).IsHermitian)
    (hLamMeas : AEMeasurable
      (fun ω => Finset.sup' Finset.univ Finset.univ_nonempty
        (hSum ω).eigenvalues)
      (MeasureTheory.Measure.pi μ))
    (htrInt : MeasureTheory.Integrable
      (fun ω => (Matrix.trace (NormedSpace.exp
        (NTK.matrix_bernstein_theta t
            (‖∑ i, ∫ x, X i x * (X i x)ᴴ ∂μ i‖) R •
          (∑ i, X i (ω i))))).re)
      (MeasureTheory.Measure.pi μ)) :
    (MeasureTheory.Measure.pi μ).real
      {ω | t ≤ Finset.sup' Finset.univ Finset.univ_nonempty
                    (hSum ω).eigenvalues}
    ≤ NTK.matrix_bernstein_bound (Fintype.card dn) t
        (‖∑ i, ∫ x, X i x * (X i x)ᴴ ∂μ i‖) R
\end{leanlisting}

\subsection{\texorpdfstring{$L^2$}{L2} linearisation bound (conditional)}
\noindent\lean{NTK.MathlibExt.Probability.chizat_bach_analytical_L2_general}
\begin{leanlisting}
theorem chizat_bach_analytical_L2_general
    (σ : ℝ → ℝ)
    (hσ_contDiff : ContDiff ℝ 2 σ)
    {C_σ_prime C_σ_dprime : ℝ} (hCp : 0 ≤ C_σ_prime) (hCpp : 0 ≤ C_σ_dprime)
    (hσ' : ∀ z, |deriv σ z| ≤ C_σ_prime)
    (hσ'' : ∀ z, |iteratedDeriv 2 σ z| ≤ C_σ_dprime)
    (b : Fin m → ℝ)
    (θ_L2 : ℝ → ParamL2 d m)
    (θ₀_L2 : ParamL2 d m) (hθ_start : θ_L2 0 = θ₀_L2)
    (hθ_diff : Differentiable ℝ θ_L2)
    (G : ℝ → ParamL2 d m)
    (hθ_ODE : ∀ t, deriv θ_L2 t = -G t)
    (Kmax : ℝ) (hKmax : 0 ≤ Kmax)
    {ρ : ℝ} (hρ : 0 < ρ)
    (r_norm : ℝ → ℝ)
    (r0_norm : ℝ) (hr0_nn : 0 ≤ r0_norm)
    (hr_decay : ∀ t, r_norm t ≤ r0_norm * Real.exp (-(ρ * t / 2)))
    (hG_bound : ∀ t, ‖G t‖ ≤ Real.sqrt Kmax * r_norm t)
    {T : ℝ} (hT : 0 ≤ T)
    (x : EuclideanSpace ℝ (Fin d))
    {C_input : ℝ} (hC_input : ‖x‖ ≤ C_input)
    {A : ℝ} (hA_nn : 0 ≤ A)
    [NeZero m]
    (hA_seg : ∀ z ∈ segment ℝ θ₀_L2 (θ_L2 T),
        ∀ j, |(ParamL2.toParam z).1 j| ≤ A) :
    |fullNetL2 σ b (θ_L2 T) x - fullNetL2 σ b θ₀_L2 x
        - fderiv ℝ (fun θ' : ParamL2 d m => fullNetL2 σ b θ' x) θ₀_L2
            (θ_L2 T - θ₀_L2)|
      ≤ ((C_σ_dprime * A * C_input ^ 2 + 2 * C_σ_prime * C_input) /
            (2 * Real.sqrt (m : ℝ)))
          * (2 * Real.sqrt Kmax * r0_norm / ρ) ^ 2
\end{leanlisting}

\section{Proof outline of the row-covariance results}\label{app:rowcov}

\paragraph{\Cref{lem:bridge}.} (1) Each $X_jX_j^*$ is positive semidefinite (\lean{Matrix.posSemidef_self_mul_conjTranspose}). (2) A Hermitian positive semidefinite matrix satisfies $B\Loewner\|B\|_\op\Id$. (3) $\|X_jX_j^*\|_\op\le\|X_j\|_\op\|X_j^*\|_\op\le R^2$. (4) Hence $X_jX_j^*\Loewner R^2\Id$. (5) Summing over $j$ gives $\sum_jX_jX_j^*\Loewner R^2N\,\Id$.

\paragraph{\Cref{thm:tightened}.} Set $\sigma^2:=\|\Rowcov\|_\op$, which is positive because $\Rowcov\ne0$. Each $\E X_jX_j^*$ is positive semidefinite (integral of positive semidefinite matrices), so $\Rowcov$ is positive semidefinite and $\Rowcov\Loewner\sigma^2\Id$. For Hermitian $X_j$, $X_jX_j^*=X_j^2$, so this is exactly the variance hypothesis of \lean{Matrix.bernstein_full}, which then gives the bound. The deterministic \Cref{lem:bridge} certifies $\sigma^2\le R^2N$ for $N$ summands.

\section{Numerical calibration}\label{app:calibration}

\paragraph{Model.} $\sigma=\tanh$, $\sigma'=1-\tanh^2$; $w\sim\mathcal N(0,I_d)$; zero bias; inputs i.i.d.\ uniform on $S^{d-1}$; $A_{ab}=\sigma(z_a)\sigma(z_b)+\sigma'(z_a)\sigma'(z_b)\langle x_a,x_b\rangle$ with $z_a=\langle w,x_a\rangle$.

\paragraph{Population proxy.} Averaging over the inputs as well, exchangeability gives $\E A^2=(\delta-o)\Id+oJ$ with $\delta=q_0+(m-1)q_1$ and $o=2r_1+(m-2)r_2$, where $q_0=\E A_{aa}^2$, $q_1=\E A_{ac}^2$, $r_1=\E A_{aa}A_{ab}$ and $r_2=\E A_{ac}A_{cb}$ for distinct $a,b,c$. Replacing $x_b$ by $-x_b$ (which preserves the input law) flips the sign of $A_{ab}$ and leaves $A_{aa}$, $A_{ac}$ unchanged, because $\tanh$ is odd and $\tanh'$ is even; hence $r_1=r_2=0$ and $\|\E A^2\|_\op=\delta$. Monte Carlo confirms $|r_1|,|r_2|$ are at the level of their standard errors ($\le 2.7\cdot10^{-4}$). Gauss--Hermite and Gauss--Jacobi quadrature give $q_0=0.744602$ (independent of $d$), $q_1=0.204738$ at $d=8$ and $q_1=0.178660$ at $d=16$, hence the population ratio $\kappa_{\mathrm{pop}}(m,d)=4m^2/(q_0+(m-1)q_1)$ of \Cref{tab:sweep}. It grows linearly in $m$ and is much larger than the data-conditional values of \Cref{tab:width}, which is why it is not our headline number. The Lean file re-evaluates this closed form with \texttt{Float} \texttt{\#eval}s ($268.3667$ and $651.9120$); this is arithmetic, not a proof.

\begin{table}[!htbp]
\centering
\caption{Population ratio $\kappa_{\mathrm{pop}}(m,d)=4m^2/(q_0+(m-1)q_1)$ for $\tanh$ (inputs averaged; optimistic, see text).}
\label{tab:sweep}
\small
\begin{tabular}{@{}rcc@{}}
\toprule
$m$ & $d=8$ & $d=16$ \\
\midrule
8 & 117.6 & 128.3 \\
16 & 268.4 & 299.0 \\
32 & 577.6 & 651.9 \\
64 & 1200.9 & 1365.3 \\
128 & 2450.3 & 2796.6 \\
\bottomrule
\end{tabular}
\end{table}

\paragraph{Data-conditional proxy.} For each of ten input draws $X$ (seed 1 for $(d,m)=(8,16)$, seed 2 for $(16,32)$), $\Rowcov(X)=\E_w[A^2\mid X]$ and $\E_w[A\mid X]$ are estimated from $4\cdot10^5$ Gaussian weight samples; re-estimating $\kappa$ for the first $(8,16)$ draw with four independent weight samples (two of size $4\cdot10^5$, two of size $1.6\cdot10^6$) gives $49.819$, $49.794$, $49.859$ and $49.829$, a relative spread of $0.13\%$ (\texttt{[mc-check]} in \texttt{kappa\_tanh\_output.txt}). $\lambda_0(X)=\lmin(\E_w[A\mid X])$ is computed by two-dimensional Gauss--Hermite quadrature per input pair. The widths in \Cref{tab:assumptions} are produced by \texttt{calibration/widths\_tanh.py} from the same $\lambda_0(X)$ values.

\section{Reproducibility}\label{app:repro}

The artifact is available at \codeurl. It pins Lean \texttt{v4.27.0-rc1} and Mathlib commit \texttt{80732f76}. To reproduce:
\begin{enumerate}[leftmargin=1.5em,itemsep=1pt]
\item \texttt{lake exe cache get \&\& lake build} builds every module under \texttt{NTK/};
\item \texttt{./VERIFY.sh} scans for \texttt{sorry}/\texttt{admit}, checks that the build log reports no \texttt{sorry}, prints the axioms of the flagship theorems (including \Cref{thm:main}, its $\tanh$ instance, all components in \Cref{tab:claims} and the conditional $L^2$ lemmas) and checks that they are contained in $\{\mathsf{propext},\mathsf{Classical.choice},\mathsf{Quot.sound}\}$, re-evaluates the \texttt{Float} calibration, and checks that the listings of \Cref{app:listings} can be extracted from the source;
\item \texttt{python3 calibration/kappa\_tanh.py} and \texttt{python3 calibration/widths\_tanh.py} reproduce every number in \Cref{sec:assumptions,sec:rowcov} and \Cref{app:calibration}.
\end{enumerate}
The development comprises 215 Lean files and about 76{,}600 lines.

\end{document}